\documentclass{article}
\pdfoutput=1
\usepackage{arxiv}

\usepackage{times}
\usepackage{soul}
\usepackage{url}
\usepackage[hidelinks]{hyperref}
\usepackage[utf8]{inputenc}
\usepackage[small]{caption}
\usepackage{graphicx}
\usepackage{amsmath}
\usepackage{amsthm}
\usepackage{booktabs}
\usepackage{algorithm}
\usepackage{algorithmic}
\usepackage{hyperref}
\usepackage[normalem]{ulem}
\usepackage{subfig}

\usepackage{amsmath,amsfonts,amssymb,amsthm}

\usepackage{booktabs}

\usepackage{xcolor}
\usepackage{mdframed}
\usepackage[T1]{fontenc}

\newcommand \W{\mathcal{W}}
\newcommand \OM{\mathcal{OM}}
\newcommand \E{\mathcal{E}}
\newcommand \C{\Gamma}
\newcommand \Z{\Theta}
\newcommand \T{\Phi}

\newcommand \PEM {PEM}
\newcommand \PEMs {$PEM$s}

\makeatletter

\title{
Modeling Perception Errors towards \\Robust Decision Making in Autonomous Vehicles
}

\author{
Andrea Piazzoni\\
ERI@N, Interdisciplinary Graduate School\\ Nanyang
Technological University\\ Singapore\\
\texttt{andrea006@ntu.edu.sg}\And Jim Cherian\\
Centre of Excellence for Testing \& Research of AVs\\
Nanyang Technological University\\ Singapore\\
\texttt{jcherian@ntu.edu.sg}\AND Martin Slavik\\
Centre of Excellence for Testing \& Research of AVs\\
Nanyang Technological University\\ Singapore\\
\texttt{martin.slavik@ntu.edu.sg}\And Justin Dauwels\\
School of Electrical and Electronic Engineering
\\ Nanyang Technological University\\Singapore\\
\texttt{jdauwels@ntu.edu.sg}
}

\begin{document}
\maketitle

\begin{abstract}
Sensing and Perception ($S\&P$) 
is a crucial component of an autonomous system (such as a robot), especially when deployed in highly dynamic environments where it is required to react to unexpected situations. This is particularly true in case of Autonomous Vehicles (AVs) driving on public roads. However, the current evaluation metrics for perception algorithms are typically designed to measure their accuracy per se and do not account for their impact on the decision making subsystem(s). This limitation does not help developers and third party evaluators to answer a critical question: \textit{is the performance of a perception subsystem sufficient for the decision making subsystem to make robust, safe decisions?}
In this paper, we propose asimulation-based methodology towards answering this question.
At the same time, we show how to analyze the impact of different kinds of sensing and perception errors on the behavior of the autonomous system.
\vspace{-2ex}
\end{abstract}

\section{Introduction}
Autonomous Vehicles (AVs)  are, arguably, going to be the first mass deployment of robots that poses a safety impact on public spaces such as roads.
It is well known that before an autonomous system can be deployed, each component must pass a set  of tests to prove
that it is capable of safely achieving its intended purpose. 
However, the Operational Design Domain (ODD) \cite{standard2018j3016} of the system may also include situations in which specific components (or subsystems) tend to exhibit diminished performance which may impact safety.
In the case of AVs, this is a major safety concern when planning for deployment on public roads.
For example, an AV may perform acceptably during daylight hours, but not very well when it gets dark.
In such situations, we could intuitively infer that the more fallible component is not the decision-making process, but rather the perception subsystem, which may not be able to correctly perceive the surroundings without sufficient lighting thus leading to undesirable AV behavior. 
On the other hand, we could also conclude that the decision making process is not robust enough to handle such specific situations \cite{Benenson2008}.
A recent (March 2018) fatal accident of an experimental AV with a jaywalking pedestrian under adverse lighting conditions is a case in point, as the investigation revealed  that the decision making was not robust against realistic perception errors \cite{NTSB2019}.
Evidently, a \textit{mereological} (part-whole) consideration is required, since neither of the subsystems is adequate or inadequate by itself; rather, their combination as a whole is necessary to obtain adequate performance.
Therefore, limiting the performance evaluation to the separate components does not address the issue of estimating whether the system will be able to operate safely under specific conditions and edge cases.
Furthermore, the current metrics designed to evaluate perception are inadequate to answer a critical question:
\textit{is the performance of a perception subsystem sufficient for the decision making subsystem to make robust, safe decisions?}

Virtual testing of AVs using simulations offers a safe and convenient way to validate safety \cite{Young2014}. However, high-fidelity models are necessary to achieve meaningful simulation results that represent the real world.  In particular, physics-based sensor simulations can generate synthetic sensor signals to directly challenge the perception; but, they are highly compute-intensive and deters the real-time execution of virtual tests under full Automated Driving System (ADS)-in-the-loop or Hardware-in-the-loop configurations.
Therefore, it is imperative to develop a feasible alternative that models the intended functionality together with the errors and uncertainty posed by the $S\&P$ subsystem of the ADS, to facilitate virtual testing.

In this paper, we provide some insights towards answering the aforesaid question and make the following contributions:
\begin{itemize}
    \item We review the state-of-the-art metrics used to measure the performance of AI-based perception algorithms, and identify their limitations in the context of decision making for an autonomous navigation task.  \item We recommend some novel directions towards building a representative Perception Error Model ($\PEM$) that can meaningfully describe the performance of the actual sensing and perception of an autonomous system.
    \item We describe an experimental setup designed to exploit the potential of $\PEM$ in a virtual (simulated) environment which offers perfect ground truth, by employing $\PEM$s to replace the actual $S\&P$ of the autonomous system architecture. By including $PEM$s, we gain the flexibility to introduce meaningful and representative perception errors while eliminating the need to generate any synthetic sensor signals.
    \item We demonstrate the usefulness of this approach as a tool to analyze how the perception capabilities of the system can impact the AV behavior, by investigating several representative urban driving scenarios based on real-life situations. The $\PEM$s considered in the experiments will also highlight the limitations of the standard evaluation metrics for perception.
\end{itemize}
\section{Decision Making Process}

\begin{figure}[t]
     \centering
     \includegraphics[width=\columnwidth]{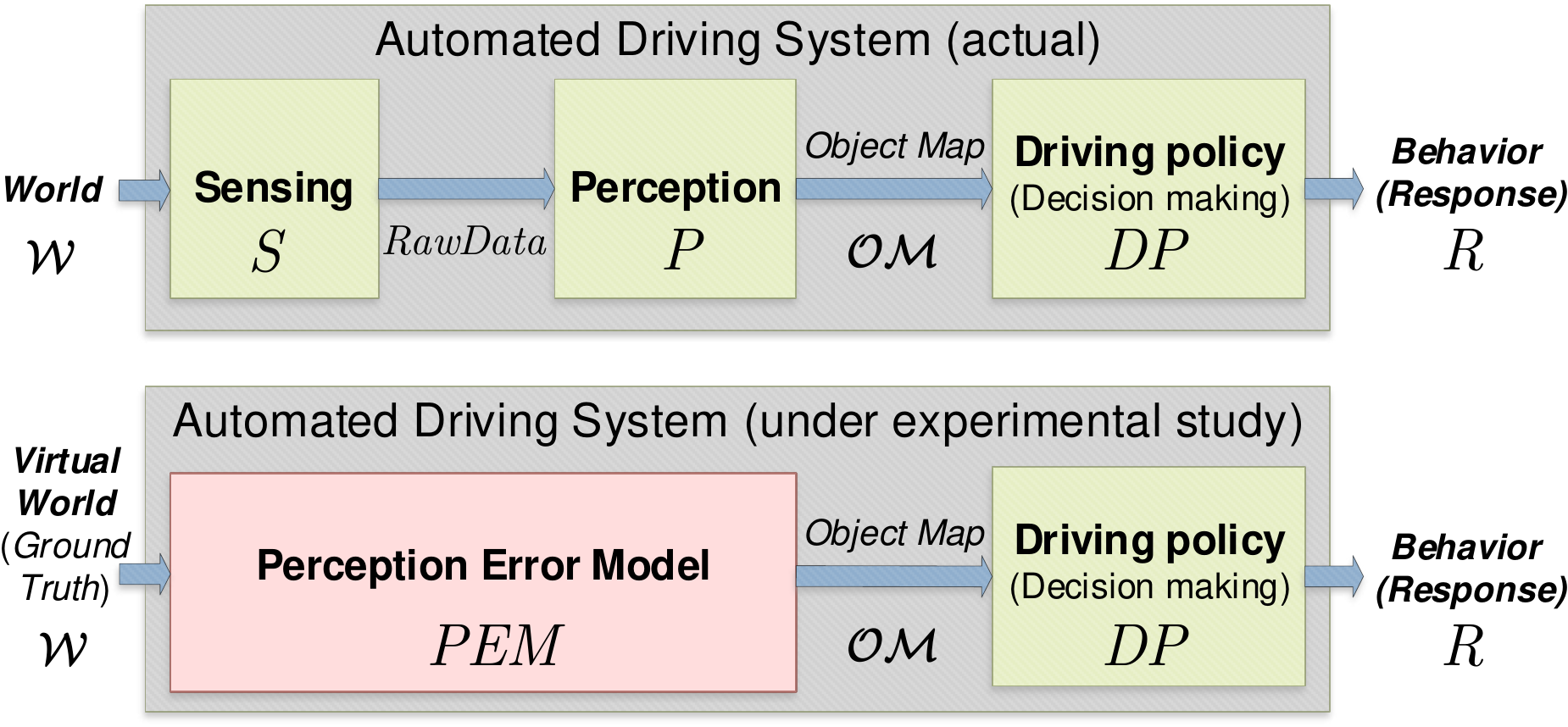}
     \caption{Simplified architecture of an ADS illustrating how $PEM$s can help to study the impact of $S\&P$ errors on decision making: (top) Actual ADS as it operates in real world (below) ADS with $PEM$ in virtual world.}
     \label{fig:simple_AV_pipeline}
\end{figure}

Most autonomous systems can be regarded as discrete-time decision making systems that operate in a continuous-time physical world.
In the context of AVs, we can term this decision making process as the Driving Policy $DP$ \cite{Shalev2017formal} that leads to a physical response, i.e. AV behavior. Although $DP$ can be hand-crafted (based on a rule book), it is tedious and less robust given the complex environment with ``surprises'' that the AV is expected to operate in. Therefore, many systems learn the art of decision-making from data using reinforcement learning \cite{Shalev2017formal}, introducing new challenges \cite{amodei2016concrete}.

While generating a robust driving policy for a robot operating in a controlled environment is generally a tractable problem, this may not be the case for autonomous vehicles operating on public roads that are shared with other traffic participants including human-driven vehicles and vulnerable road users (such as pedestrians or cyclists). Therefore, it is very important to ensure that despite the complexity of the ODD, the driving policy is robust enough to generate an appropriate AV behavior which is safe as well as comfortable to the passengers in real time.
This is arguably a complex dynamic spatio-temporal optimization problem, 
wherein the constraints possess high \textit{aleatoric} as well as \textit{epistemic} uncertainty \cite{McAllister2017}.
Therefore, the AV research community (including industry, academia and regulators) tries to select scenarios to generate appropriate test cases, and relevant safety metrics that are measurable, objective and robust. Typical safety metrics used include safety clearance distances (between AV and other traffic participants), maximum and minimum limits on AV speed, acceleration and deceleration and many others. Nevertheless, a simple and practical metric is the \mbox{clearance} distance, both in temporal and spatial domains.

\section{Error Model}

The surrounding environment can be summarized by 3 elements: the map, the ego-vehicle localization, and the other road users or obstacles.
In this paper, we focus on the detection of obstacles and other road users. These are described in the Object Map $\OM = \{o_1, \ldots ,o_m\}$, the set of the $m$ perceived objects that the $S\&P$ system provides for each frame by observing the surrounding world $\W = \{w_1, \ldots ,w_n\}$. The world $\W$ represents the set of $n$ detectable objects (a.k.a. the Ground Truth).
The $\OM$ is then analyzed and through the Driving Policy $DP$, a response $R$ is generated.

To describe each object $w \in \W$, we adopt the same notation as for objects $o \in \OM$ where $o = (\mathbf{X}, \mathbf{C})$:
\begin{itemize}
    \item \textbf{Pose}: the (6-9)DoF pose $\mathbf{X}$ of an object $o$,  represented as a vector of 9 parameters (a 3D bounding box),
    \begin{equation} 
        \mathbf{X}= \big(\text{\small{\textsf{position}, \textsf{rotation}, \textsf{dimensions}}}\big).
    \end{equation}
    This includes 3 parameters each for {\small\textsf{position}} (x, y, z),  {\small\textsf{rotation}} (yaw, pitch, roll), and {\small\textsf{dimensions}} (length, width, height).
    Some parameters such as pitch, roll, or height may be dismissed in specific road traffic environments.
    \item \textbf{Class}: the class $\mathbf{C}$ of the object, 
    \begin{equation}
        \mathbf{C} \in \{Vehicle, Pedestrian\}.
    \end{equation}
    Depending on the system, it can vary from a simple distinction between vehicles and pedestrians, to a finer classification discriminating between cars, bikes, trucks, etc.
    \item \textbf{Additional Parameters}: vector $\mathbf{X}$ can be extended to include any other relevant object parameters such as \mbox{velocity}, turning indicator status, or age (for pedestrians), based on the system under consideration and $\mathbf{C}$.
\end{itemize}

In \autoref{fig:simple_AV_pipeline}, we note $RawData = S(\W)$ and $\OM = P(RawData)$. We can observe that: 
\begin{equation}\label{eq:W+error}
        \OM = P(S(\W)) = S\&P(\W) = \W + \E,
\end{equation}
where $\E$ is the error between $\OM$ and $\W$.
The response $R$ is what determines the \textit{behavior} of the vehicle and therefore, the overall safety and performance of the autonomous system:
\begin{equation}\label{eq:Dp+error}
        R = DP(\OM) = DP(\W + \E).
\end{equation}

The task of assembling the $\OM$ requires to address both classification and regression problems, and has its roots in the object detection task in the Computer Vision (CV) field.

\subsection{Evaluation Metrics - State of the Art}

The error $\E$ includes predominantly 4 kinds of error. Let $o_i$ be an object perceived corresponding to $w_j$:
\begin{itemize}
    \item \textbf{False negative}: $ o_i \notin \OM $;
    \item \textbf{False positive}: $w_j \notin \W$;
    
    \item \textbf{Misclassification}: $\mathbf{C}_{o_i}\neq \mathbf{C}_{w_j}$; 
    \item \textbf{Parameters errors}: $\mathbf{X}_{o_i} - \mathbf{X}_{w_j} \neq \mathbf{0}$.
\end{itemize}

All these kinds of errors can be individually observed, statistically measured, and studied by comparing $\W$ and the $\OM$ produced by the $S\&P$.
Given \autoref{eq:W+error}, the task of analyzing and describing the error is an extension of the task of \textit{measuring} the error of a perception subsystem. In fact, many metrics have been developed for the task of object detection from the CV field.
In the field of AVs, many benchmarks on public datasets \cite{Geiger2012,Huang2018a,nuscenes2019} explore variations of these metrics. 

Intersection over Union (IoU) and Mean Average Precision (mAP) are popular metrics for assessing CV algorithms for generic object detection tasks \cite{Everingham2010,Cordts2016Cityscapes}.
Similarly, Multiple Object Tracking Accuracy (MOTA) and Multiple Object Tracking Precision (MOTP) are common metrics for tracking evaluation \cite{Bernardin2008}.
All of these metrics require an \textit{a priori} definition of a threshold in order to discriminate between a true positive and a false positive. 
For example, in \cite{Geiger2012} the authors consider IoU $\geq$ 0.7 for the correct detection of a car, or IoU $\geq$ 0.5 for a pedestrian.

\subsubsection{Evaluation Metrics - Critical Issues} \label{ssec:criticalities}
In the deployment of AI-based systems, it is not rare that accuracy is not necessarily the best metric to measure their capabilities \cite{PadovaniFL19bardo}. 
While it is not debatable that having a perfect score on accuracy-based metrics is the final goal (i.e., the perception perfectly overlaps with the ground truth, implying $\OM = \W$), these metrics were not designed to consider $DP$.  
Hence, they do not provide a model that is adequate enough to study \autoref{eq:Dp+error}.
This is because the use of a single metric would hide the specifics of the type of error causing perturbations in the measurement.

\begin{figure}
     \centering
     \includegraphics[width=\columnwidth]{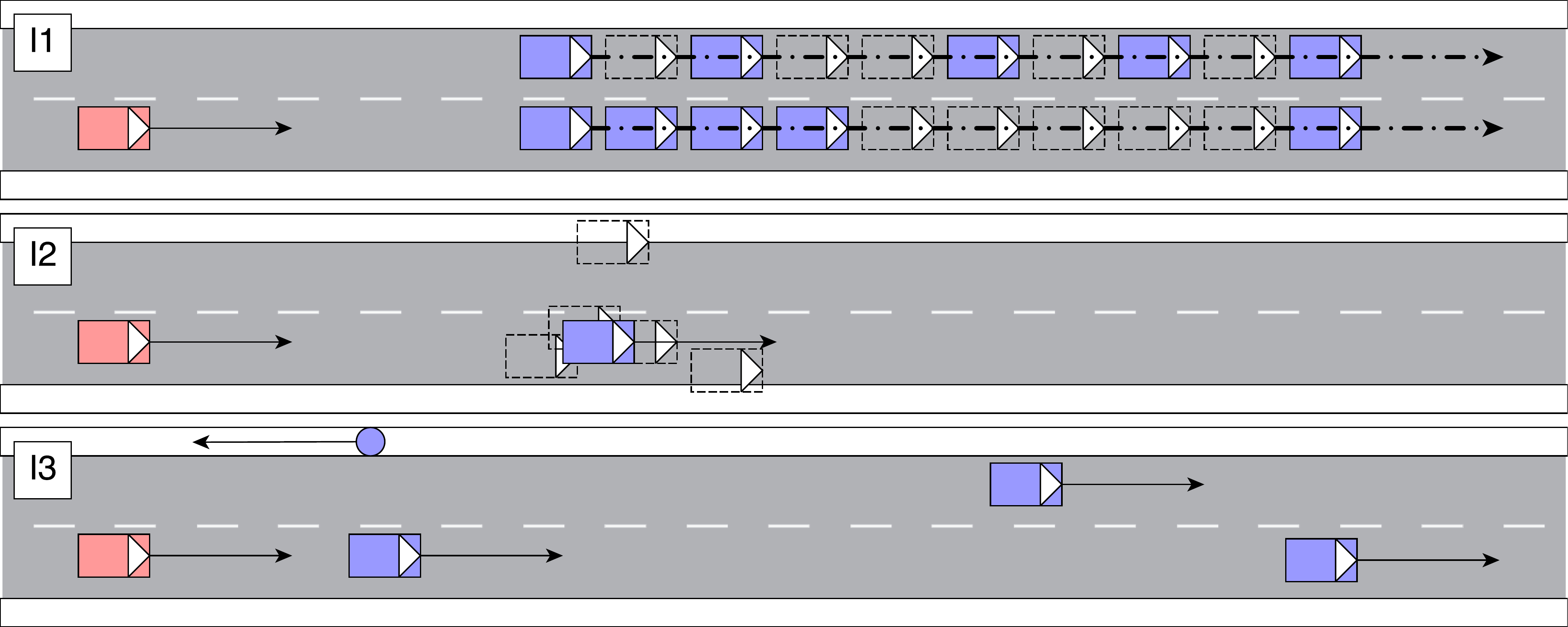}
     \caption{Illustration of the critical issues I1, I2, I3.\\
     I1: Temporal considerations: short vs. long non detection intervals.\\
     I2: Overlap Sensitivity: how sensitive is $DP$ to spatial error?\\
     I3: Relevance of the objects: which ones are active constraints?}
     \label{fig:issues}
\end{figure}

In particular, we identify 3 critical areas for analyzing the response $R$ (\autoref{fig:issues}), but are out of the scope of CV metrics:

\begin{itemize}
    \item \textbf{I1: Temporal relevance}: if the system is deployed in a highly dynamic environment, the worst-case error (e.g., losing track of an object for longer intervals) may be more relevant than the average error for same duration.
    \item \textbf{I2: Overlap sensitivity}: The spatial error associated to each object is definitely important. However, considering the bounding box overlap alone may not be sufficient to gauge the quality of the response provided by $DP$.
    \item \textbf{I3: Relevance of the objects}: 
    Generic CV tasks do not usually associate a weight to each object, as the context may not be considered. However, for an AV in a well-defined ODD, the metrics should judge the relevance of objects considering the context and dynamics (refer I1).

\end{itemize}

For a more abstract understanding, we should ask: \textit{If the system response $R$ provides the desired outcome, such as avoiding a collision, does it really matter if the $\OM$ had significant errors?}
E.g., if the AV brakes to avoid a perceived pedestrian, how much does it matter if the object was actually a cyclist?
In this case, how to quantify the relevance of a specific error?
There is no straightforward answer to this, since a major classification error could also cause the AV to respond in an unacceptable and/or unsafe manner.

\subsection{Error Modeling Considerations}
\label{ssec:err_model_considerations}
To better understand how the error manifests itself and to subsequently analyze the performance of the $S\&P$, we must first understand the causes of the error.

\paragraph{Positional aspects:} Our first observation is that the quality of $S\&P$ is influenced by the \textbf{relative position} of $w$ w.r.t. the ego-vehicle, for 2 reasons \cite{rosique2019systematic}:
\begin{itemize}
    \item \textbf{Distance}: Performance of all sensors degrades at longer distances. E.g, a more distant object will be captured by fewer pixels by a camera and by fewer LiDAR points.
    \item \textbf{Field of View (FoV)}: Sensors cover different areas around the ego-vehicle. An object that is positioned in an area covered by multiple sensors could be detected with greater accuracy than an object located in an area covered by few, or weaker, sensors.
\end{itemize}

\paragraph{Parameter inter-dependencies:} The second aspect is that the values of any of the object parameters $\mathbf{X, C}$ can, by themselves, affect the error associated with other parameters of $\mathbf{X, C}$ \cite{Hoiem2012}. For example, a larger \textit{size} of an object makes it more likely to be seen at greater distances, whereas the object class $\mathbf{C}$ may limit the error on the size estimation. Some parameters are also not described in the $\OM$ since they are not directly relevant for $DP$, such as the color or the material of the object \cite{rosique2019systematic}. For example, dark/non-reflective surfaces or metallic artifacts may degrade the quality of $\OM$ when $S$ is primarily based on \mbox{LiDAR} or Radar respectively.

\paragraph{Temporal aspects:} As a third observation, we can consider the \textit{temporal} aspects of the system, since the $DP$ deals with a sequence of detections, a dynamical system and environment. The overall error of the system can change over time, due to shifting light conditions (e.g., sun blinding, shadows), algorithm uncertainties or even any interference at the level of individual sensors,  and should be modeled appropriately.
Errors evolve over time and hence should be viewed as time series and be modeled by dynamical models \cite{Mitra2018}. 

\subsection{
Perception Error Model}
In this paper, we propose a Perception Error Model ($\PEM$) that comprises both the sensing subsystem $S$ and a perception subsystem $P$, approximating their function:
    \begin{equation}
    PEM(\W) \approx S\&P(\W) = \OM = \W + \E. 
    \end{equation}
We propose the following abstraction: 
\begin{equation}\label{eq:pem}
      PEM = \{  \T, \Z, \C \},
\end{equation}

where each component is defined as follows:
\begin{itemize}
    \item $\T$: temporal and statistical description of the perception error in function of $\mathbf{X}$;  
    \item $\Z$: a zone-based spatial description of the $S\&P$ error distribution around the AV considering the coverage by sensors (as illustrated in \autoref{sensors_illustration}), addressing the positional aspects, viz., the FoV and Distance problems.
    
    \item $\C$: a description of environmental conditions affecting the error, e.g., a system deployed on the road can be conditioned by the light intensity or fog density, which can be modeled as continuous variables in $\T$. Alternatively, one could choose to discretize them and provide distinct $\T$ for each value (e.g., $\T_\text{daylight}$, $\T_\text{night}$).
\end{itemize}

    \subsubsection{Zones-based approach for $\Z$}
\begin{figure}[bt]
         \centering
         \subfloat[a][Camera]{\includegraphics[width=.3\columnwidth]{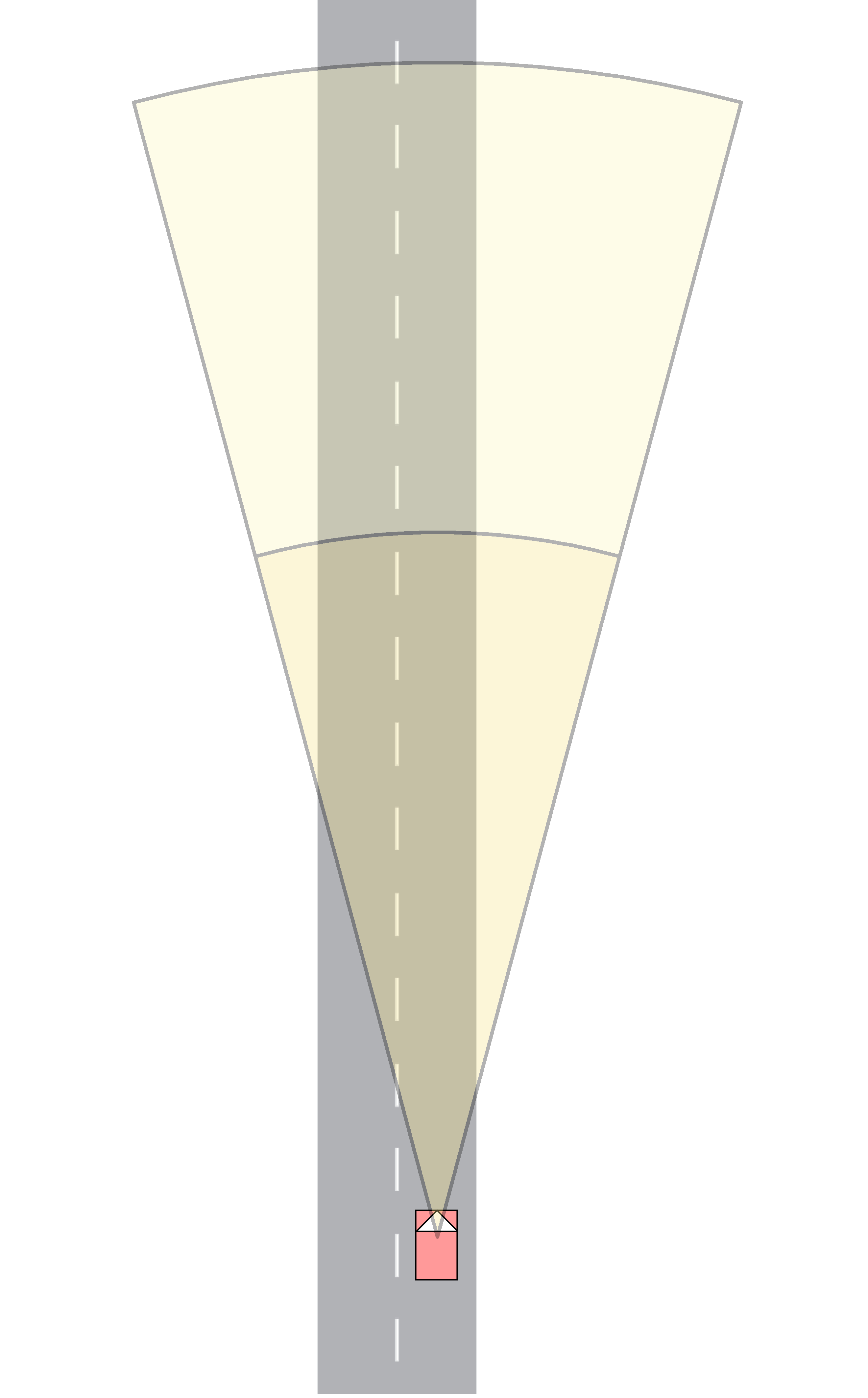}\label{fig:sensor1}}
         \subfloat[b][LiDAR]{\includegraphics[width=.3\columnwidth]{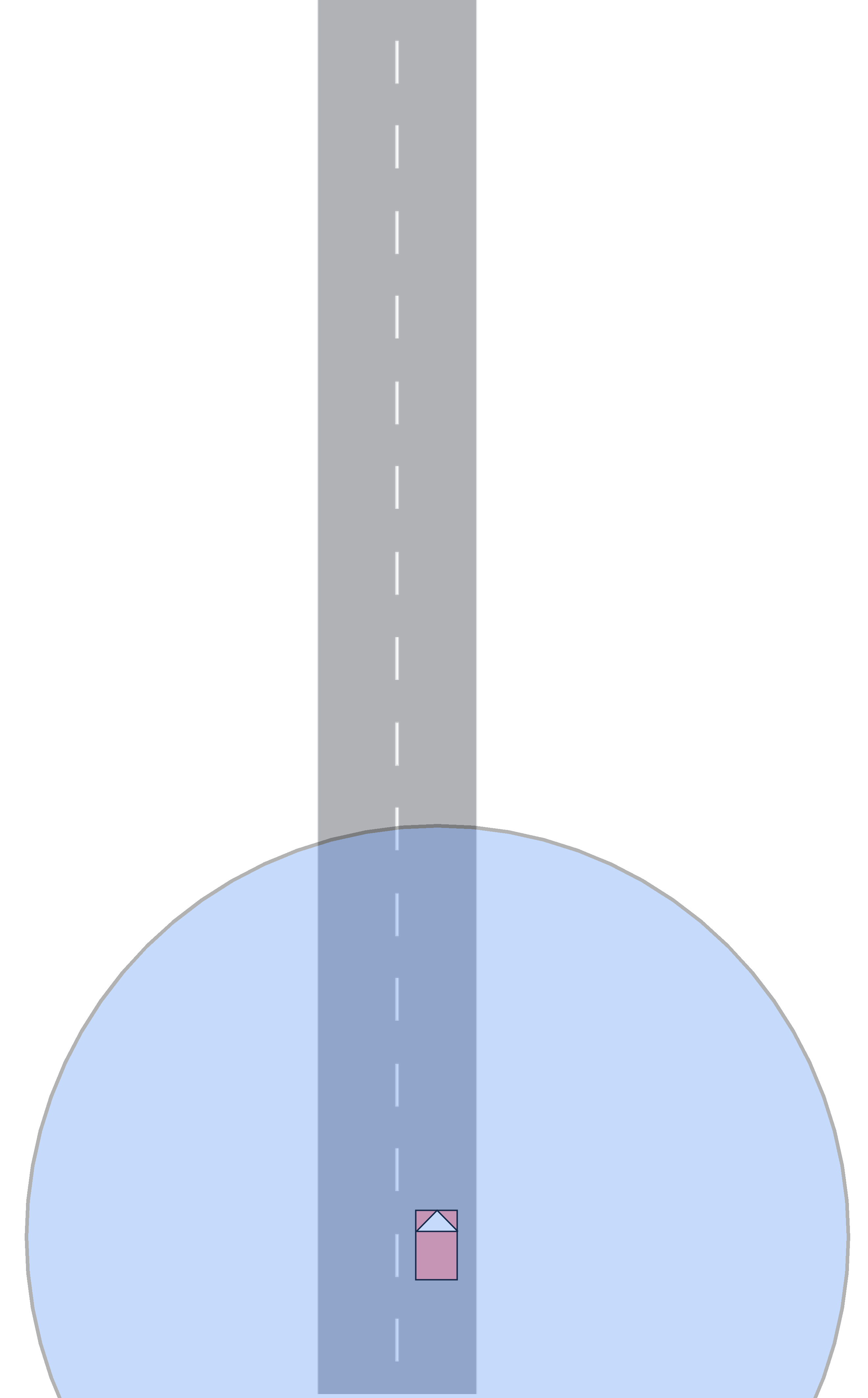}\label{fig:sensor2}}
         \subfloat[c][Overlap]{\includegraphics[width=.3\columnwidth]{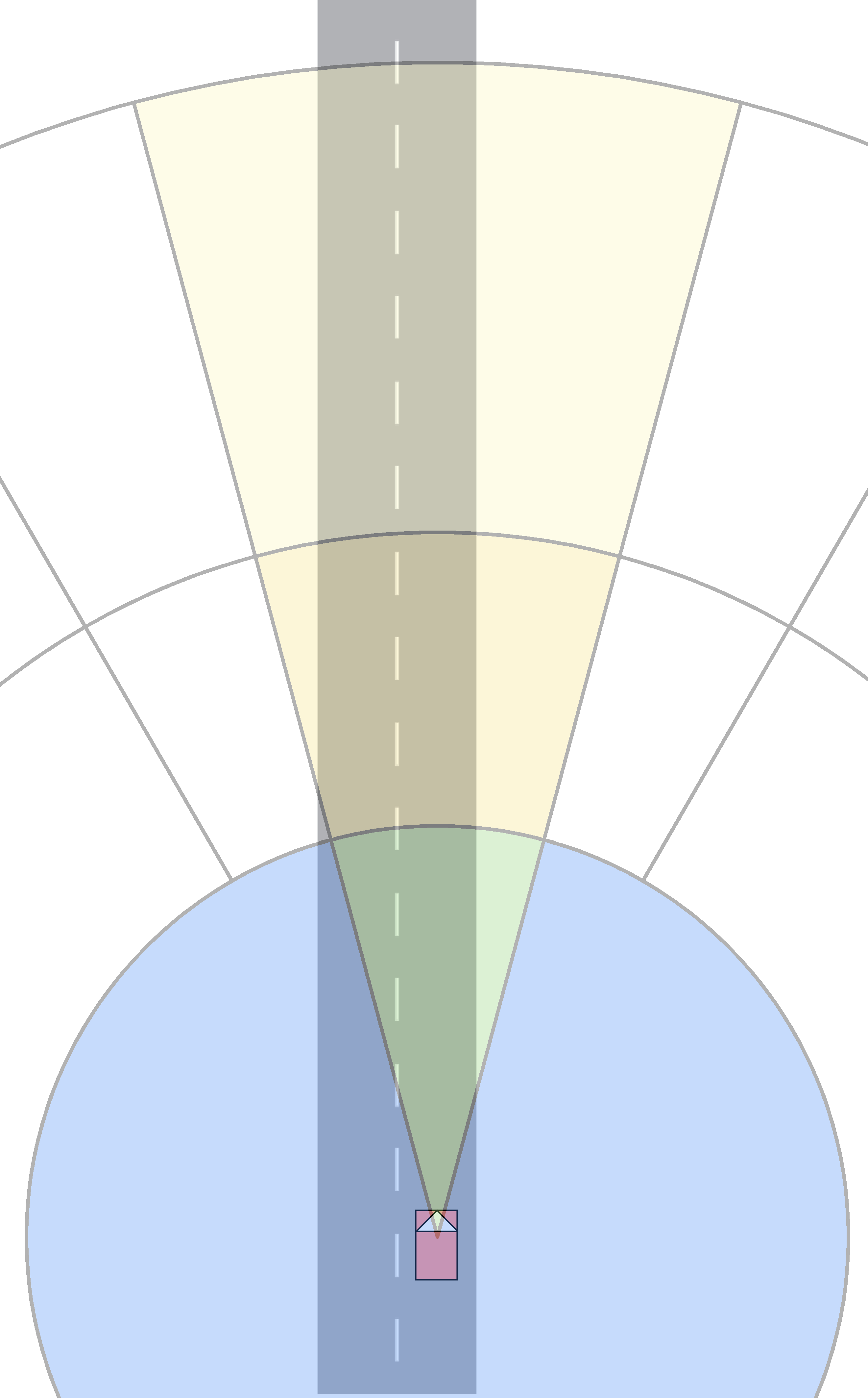}\label{fig:sensor3}}
         
         \caption{Example of a zone-based partitioning of the $S\&P$ errors. (a) Simple camera FoV, divided into 2 zones based on its range; (b) LiDAR, with 1 zone; (c) Multiple sensors, leading to an overlap region (or regions) where objects can be detected by both sensors. The perception error will depend on how the signals are fused.}
         \label{sensors_illustration}
    \end{figure}

    As illustrated in \autoref{sensors_illustration}, we propose to address the positional aspects of error by representing the perception error in different \textit{zones}.
    The FoV problem is easily solved, by dedicating a zone to the overlap of a specific set of sensors.
    The distance problem instead is already considered in some CV benchmarks \cite{Cordts2016Cityscapes,waymo_open_dataset}. The common solution is to discretize the distance in different ranges, breaking down the metrics into different regions.
    Our \textit{zones} approach is, in fact, an extension of that approach; while the zones can also be determined by distance thresholds, we also make the entire approach \textit{sensor-agnostic}.
    Furthermore, we can apply the \textit{zones} approach to study the contextual \textit{relevance} of the objects for a given driving scenario and a planned manoeuvre. Dedicated models for each zone allows to better understand which are the critical areas of the surroundings.

    \subsubsection{Key considerations for $PEM$}
    If $\T$ is designed to simply return each object $w$ without any alteration in its parameters, the model is replicating a perfect $S\&P$ system that is able to detect the ground truth. 
    More interestingly, the model can also be designed to not return objects in specific zones $\in \Theta$, thus replicating cases of non-visibility such as blind spots (i.e. the object is not within the range of any sensor, or is occluded \cite{suchan2019}).

    Considering the above, we propose that designing a $\PEM$ is, without loss of generality, a \textit{regression} task, where the goal is to learn the rules and parameters which describe the difference ($\E$) between $\W$ and the $\OM$ generated by the $S\&P$ subsystem (\autoref{eq:W+error}), formalized as $\Phi$, $\Theta$, and $\Gamma$.
    
    Despite having similarities with a typical procedure for computation of evaluation metrics (i.e. comparing perception output to the ground truth), this task is more complex. As described in Section \ref{ssec:err_model_considerations}, it involves analyzing influence of spatio-temporal dependencies, and object-specific parameter inter-dependencies (covariances) which are relatively under-explored fields.
    Such aspects are mainly conditioned by the choice and configuration of sensors and perception algorithms. Thus, academic studies focus mostly on a generalized performance evaluation.

    To motivate research in this direction, in the next section we focus on showing how a richer and more descriptive $\PEM$ can serve to address some of the issues in the evaluation of both the $S\&P$ and $DP$.
    Such an evaluation process is not only crucial from a regulatory perspective, but can also facilitate the system development life-cycle; it can guide developers in choosing the sensors, training the perception models, as well as identifying weaknesses in the $DP$.
\section{Experimental Setup}

\begin{figure}[t]
\centering
\includegraphics[width=\columnwidth]{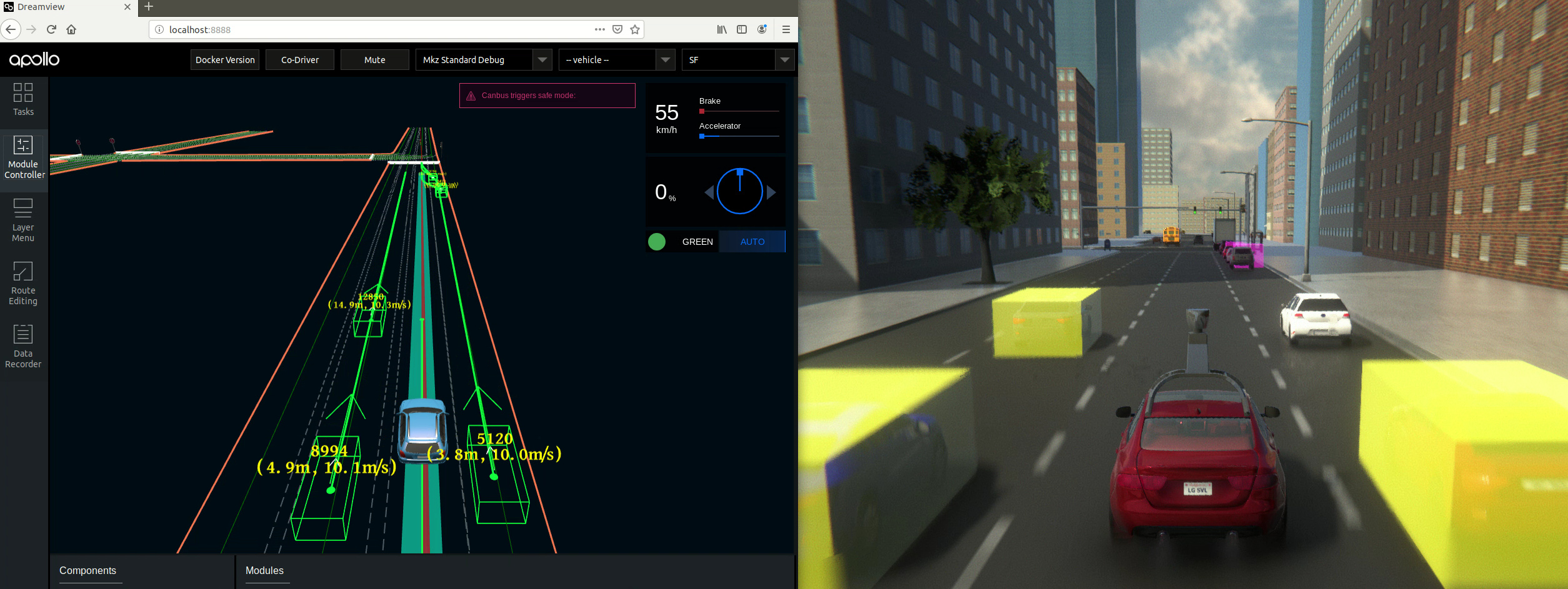}
\caption{Screenshot of the co-simulation in a generic urban driving situation. Bounding boxes (yellow/purple) of $PEM$-based objects $\OM$ rendered by LGSVL (right) are consistent with what Apollo sees (left), even undetected objects. 
}\label{fig:cosim}
\end{figure}

\begin{figure*}[t!]
     \centering
     \subfloat[Illustration of 2 scenarios in our experiments (TC1-3, TC4-5).]{\includegraphics[width=0.95\columnwidth]{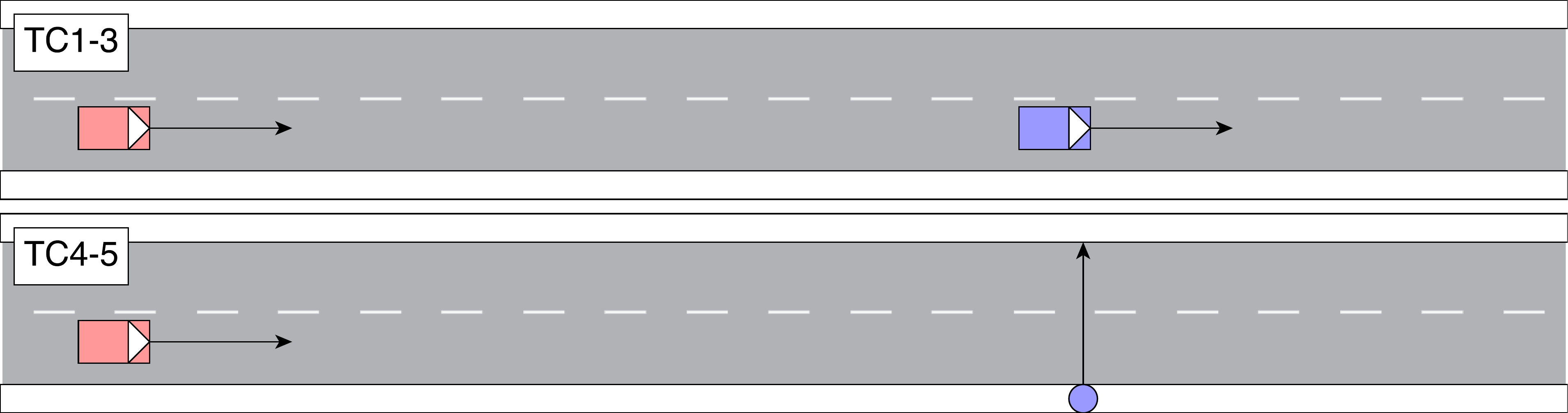}}
     \hspace{4mm}
     \subfloat[Following another vehicle.]{\includegraphics[width=0.48\columnwidth]{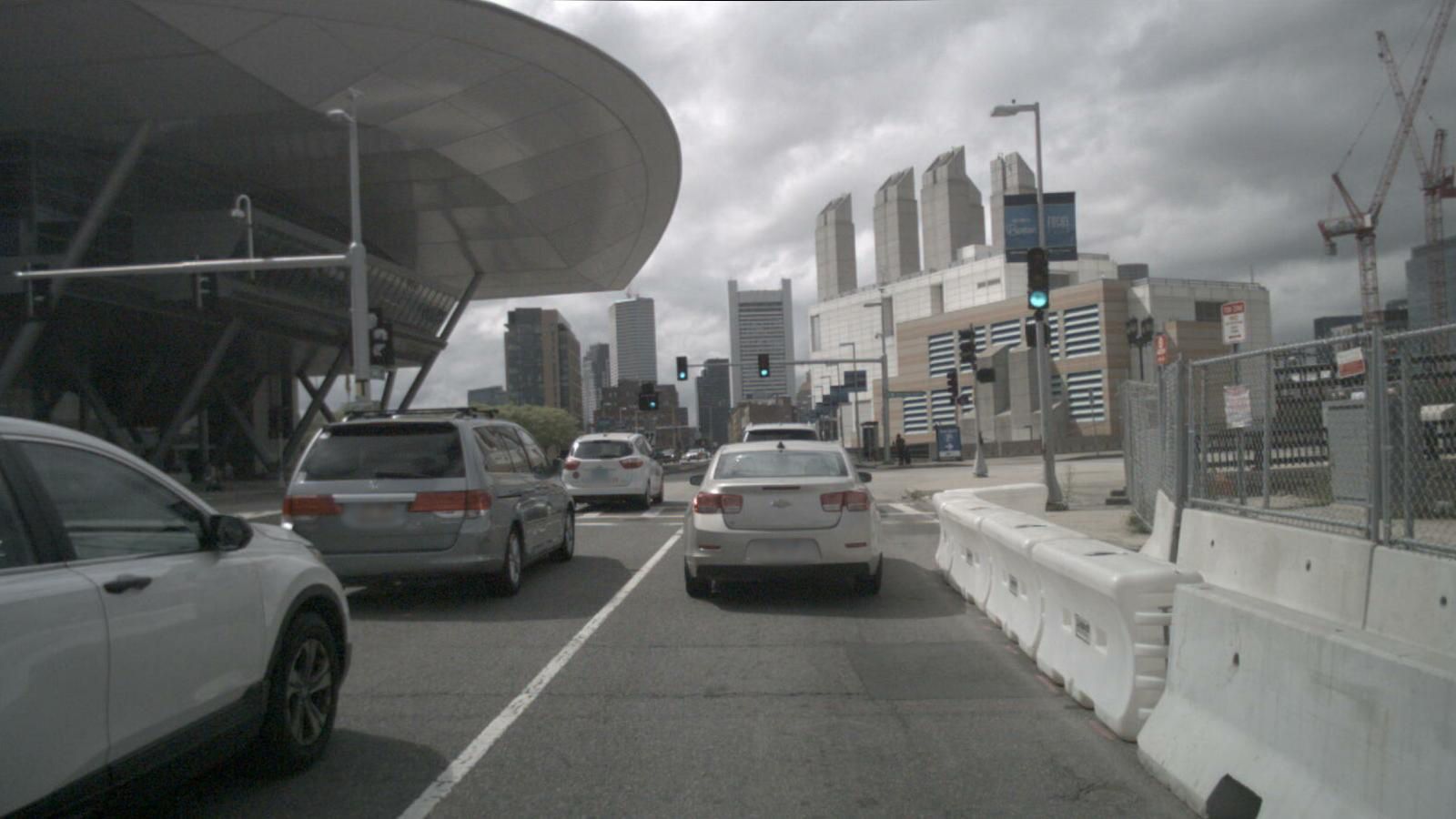}}
   \subfloat[Pedestrian on an urban road.]{ \includegraphics[width=0.48\columnwidth]{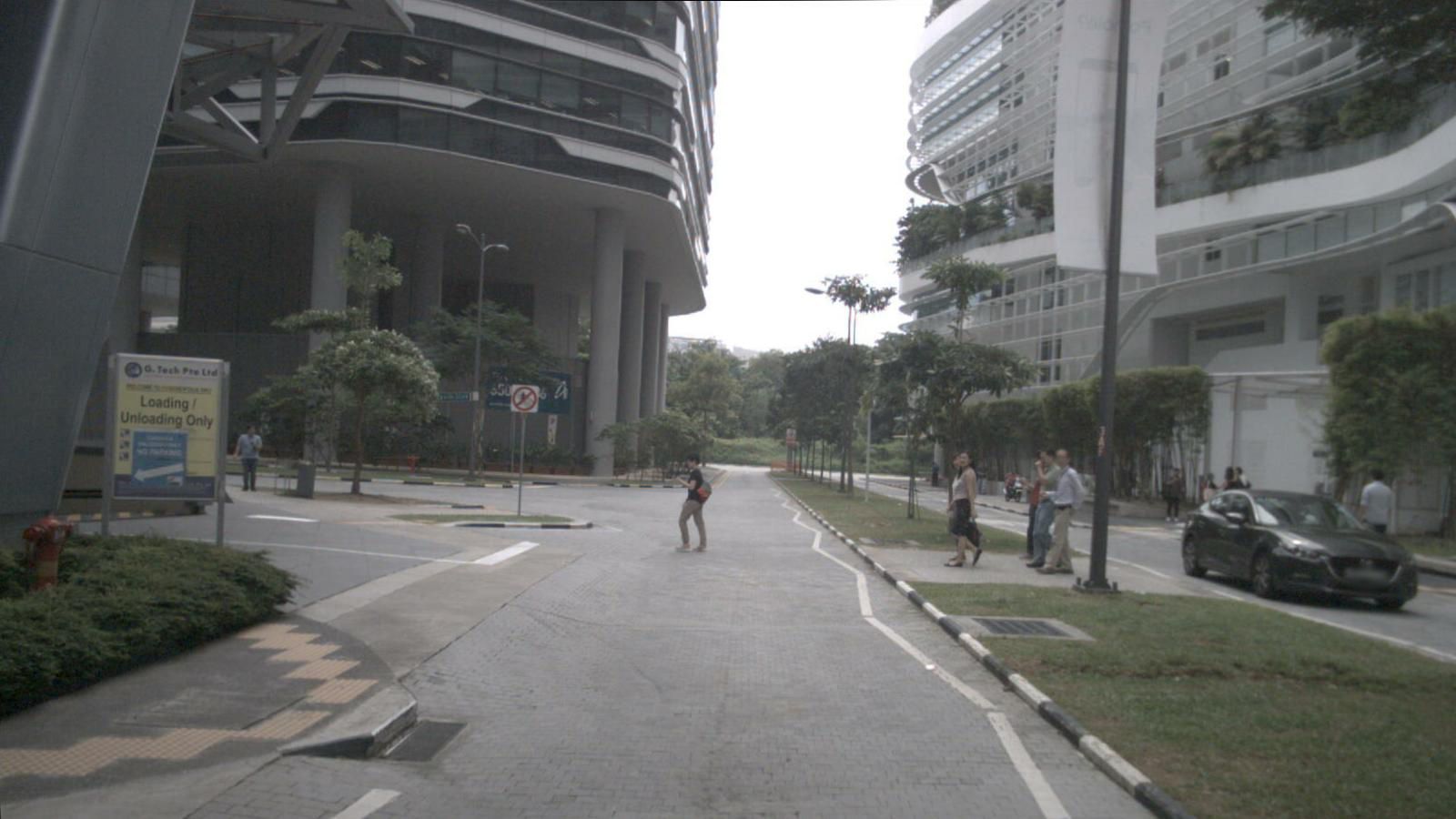}}
    \vspace{-1mm}
    \caption{Scenarios used in experiments: (a) a representative illustration, (b,c) instances from the nuScenes dataset \protect\cite{nuscenes2019}.}
    \vspace{-2mm}
     \label{fig:scenarios}
\end{figure*}
\begin{figure}
\centering
\includegraphics[width=\columnwidth]{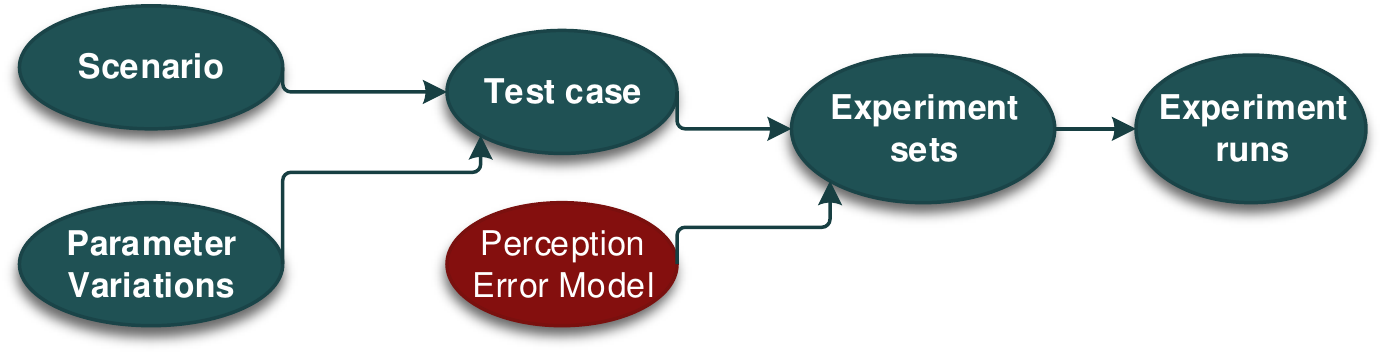}
\caption{Functional relationship between scenarios, parameter variants, test cases and error models in our experiments.}\label{fig:exp}
\end{figure}

In this section, we describe the software tools and the experiments conducted to highlight how different kinds of error do (or do not) affect the response $R$, thus allowing us to observe how a specific $PEM$ can influence $R$.
Furthermore, we describe different $PEM$ variants that serve to demonstrate some of the {critical issues} discussed in the earlier sections.
In this paper, we focus on $\T$, i.e., the temporal and statistical description of $S\&P$ error, and specific statistics related to standard evaluation metrics.
The experiments we designed require a driving simulator and an ADS.
To this end, we chose open source tools, namely LGSVL simulator \cite{LGSVL} and Apollo 3.5 \cite{Fan2018}.
LGSVL simulator is based on the Unity Engine and maintains a reliable bridge between Unity framework and the CyberRT middleware relied on in Apollo 3.5, thus enabling co-simulation (see \autoref{fig:cosim}). 
We developed python scripts to implement different scenarios,  to automate the tests, configure the simulation environment and the actors in a deterministic manner,
and to log the results.

To facilitate our experiments, we adapted these tools so that we could include the $\PEM$ in the loop. 
To this end, we bypassed the built-in $S\&P$ subsystem in Apollo. Firstly, instead of processing (synthetic) raw sensor data, we adapted Apollo to directly read the $\OM$ from a new special-purpose CyberRT topic. Secondly, we defined a new sensor in LGSVL simulator that upon observing $\W$, generates $\OM$ by applying the specific $\PEM$ (see \autoref{eq:pem}) configured for the experiment, and then publishes $\OM$ on the new CyberRT topic that the decision making part of Apollo could read from.

\subsection{List of Scenario-based Experiments}
In order to study the influence of error models on the AV behavior, we generated a set of experiments following the scheme depicted in \autoref{fig:exp}. In particular, we defined a set of relevant driving scenarios (see \autoref{fig:scenarios}), configured their parameter variations to get concrete test cases, and tested them with different $PEM$s to form actual experiments that are executed multiple times (at least 30 runs each, to account for randomness involved in our $\PEM$s).

\paragraph{Scenario 1 (Test cases TC1-3):} involves an AV driving on a straight road, approaching a traffic vehicle and then following it until they reach a red traffic light.
For each test case, each traffic vehicle was set to drive at one of 3 different average speeds, viz. 7, 10, and 15 m/s.
To challenge the $DP$, 
we applied $PEM$s that can correctly detect an object ($o_i = w_j$) but randomly fail to include it in $\OM$ for some frames (similar to tracking loss or sporadic non-detections).
This allows us to study the critical issue of temporal relevance (see I1).\\
\textbf{Implementation TC1-3:} We model the False negative errors by means of Markov chains with two states.
We tested different values of the parameters \textit{steady state probability }$\in [0.0,1.0]$ and \textit{mean sojourn time} $\in (0.0s,10s]$ (average time spent in a state before changing) so as to generate non-detection intervals of varying duration.

\paragraph{Scenario 2 (Test cases TC4-5):} this is defined by the presence of a pedestrian in 2 different situations: standing in the middle of the road, or jaywalking.
For these TCs, we applied $\PEM$s that generate different \textbf{positional errors}, with the intention of studying the impact of critical issue I2.\\
\textbf{Implementation TC4, TC5a:} Gaussian White Noise with varying standard deviation $\sigma$, applied to the relative position of $w$ w.r.t. the AV, in polar coordinates: 
\begin{itemize}
    \item multiplicative noise on radius $d$ as $\sigma_d \in [0\%,12\%]$;
    \item additive noise on azimuth $\theta$ as $\sigma_\theta \in [0^{\circ},1.5^{\circ}$].
\end{itemize}

We then apply additional \PEMs\ to TC5, so as to replicate the failures that led to a recent AV accident \cite{NTSB2019}.\\
\textbf{Implementation TC5b}: Perfect detection at each frame, but with a tracking loss probability $p_{tl} \in [0,1]$ for the previously detected obstacles. This can result in considering the current detection as a new obstacle, which can hinder the computation of obstacle velocity and lead to unsafe behavior.

\section{Experimental Results}
In this section, we show the scope of analysis afforded by our experimental setup.
Our analysis focuses on the \textit{behavior} of AVs in particular, although the methodology can also be applied elsewhere. 
For ease of understanding, we show several representative examples from our experimentation. These examples serve to demonstrate the effectiveness of our approach in analyzing how the $PEM$s can impact the behavior, and thereby taking simple safety metrics under consideration.

\newcommand\sizescatter{.17\textwidth}
\newcommand{\rulesep}{\unskip\ \vrule height 26mm width .01mm}
\begin{figure*}[]
     \centering
     \subfloat[][TC(1-3)]{\includegraphics[width=\sizescatter]{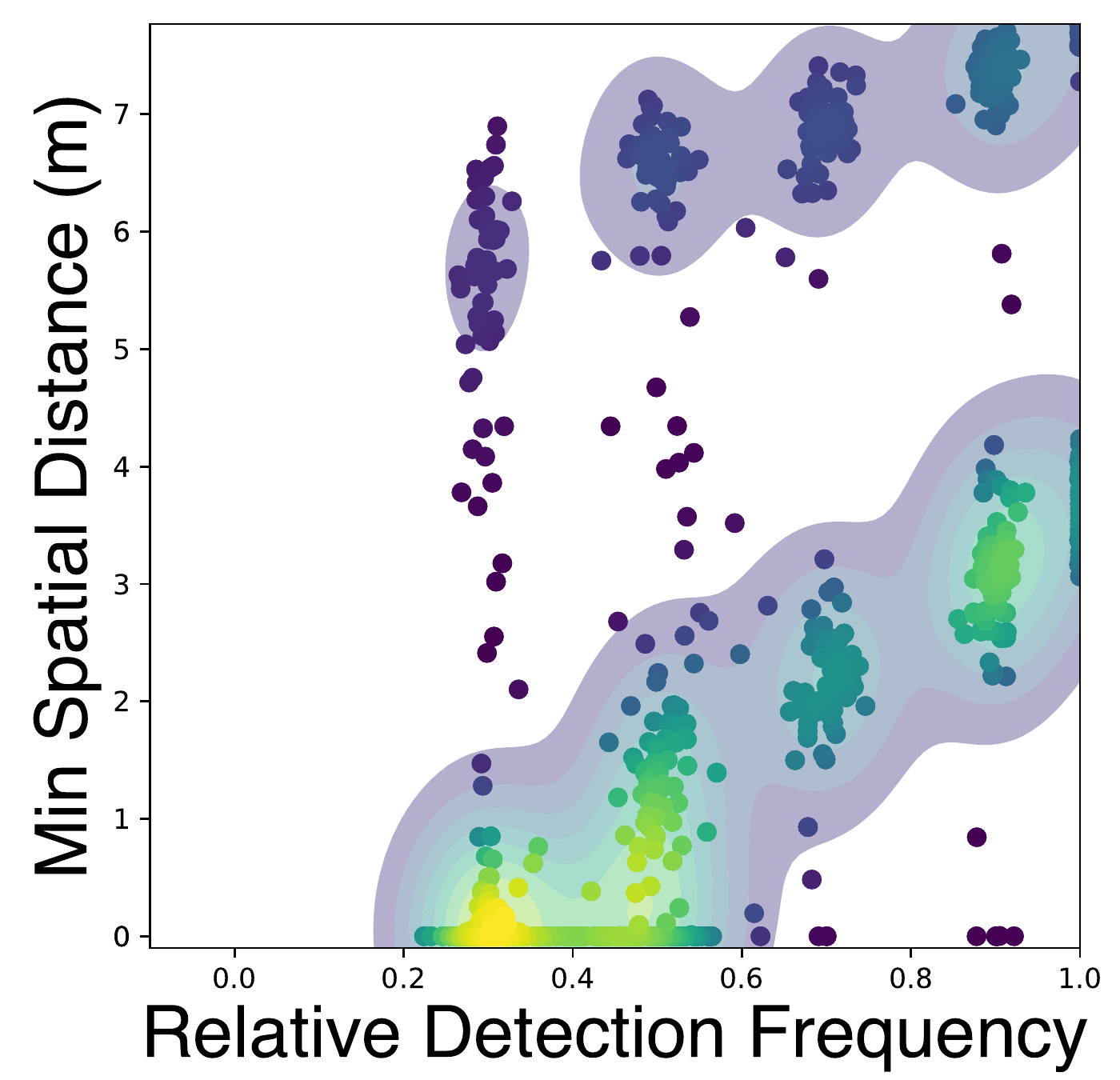}\label{fig:sp_a}}
      \hspace{1mm}
     \subfloat[][TC(1-3)]{\includegraphics[width=\sizescatter]{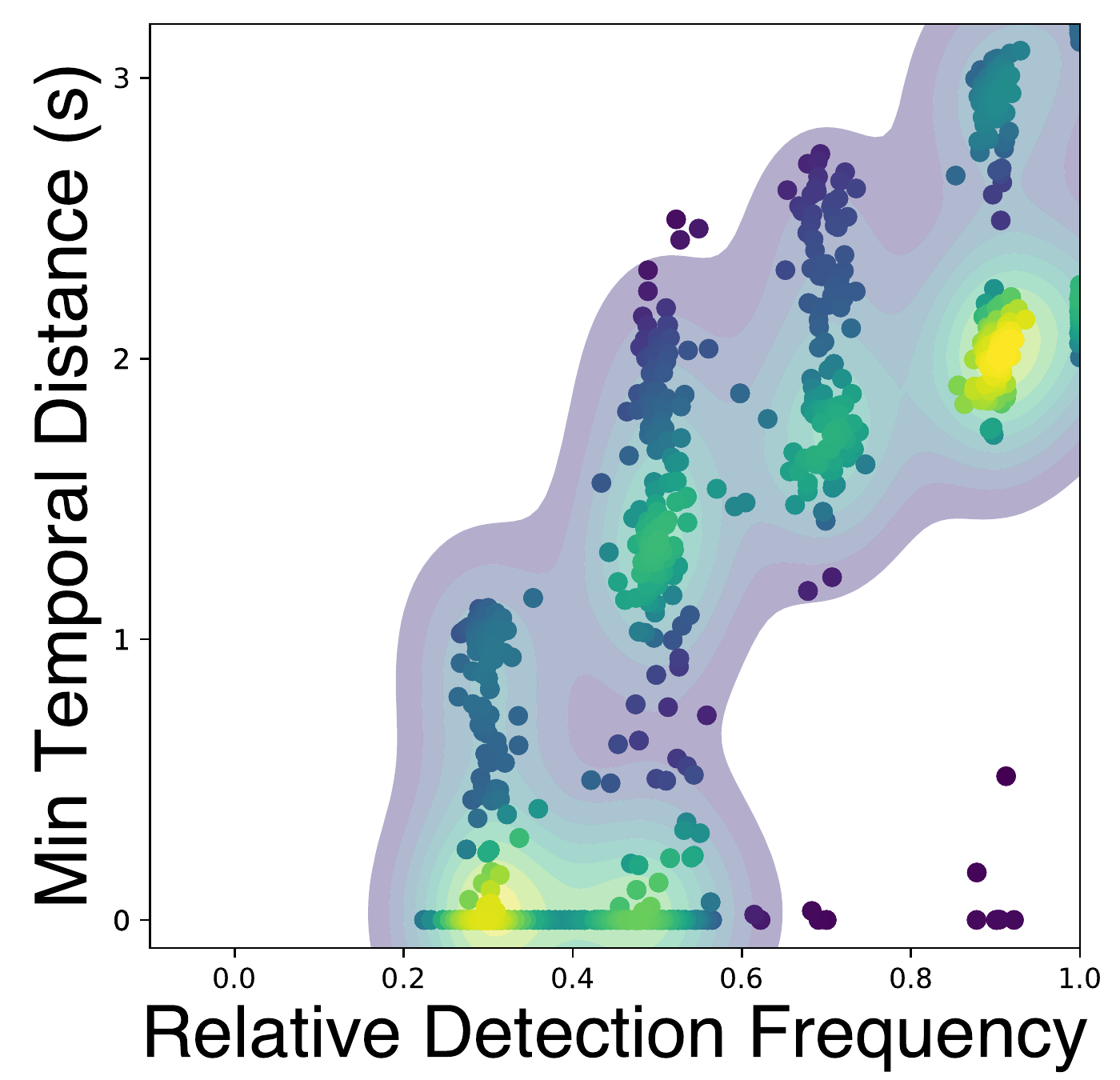}\label{fig:sp_b}}
      \hspace{1mm}
      \subfloat[][TC(1-3)]{\includegraphics[width=\sizescatter]{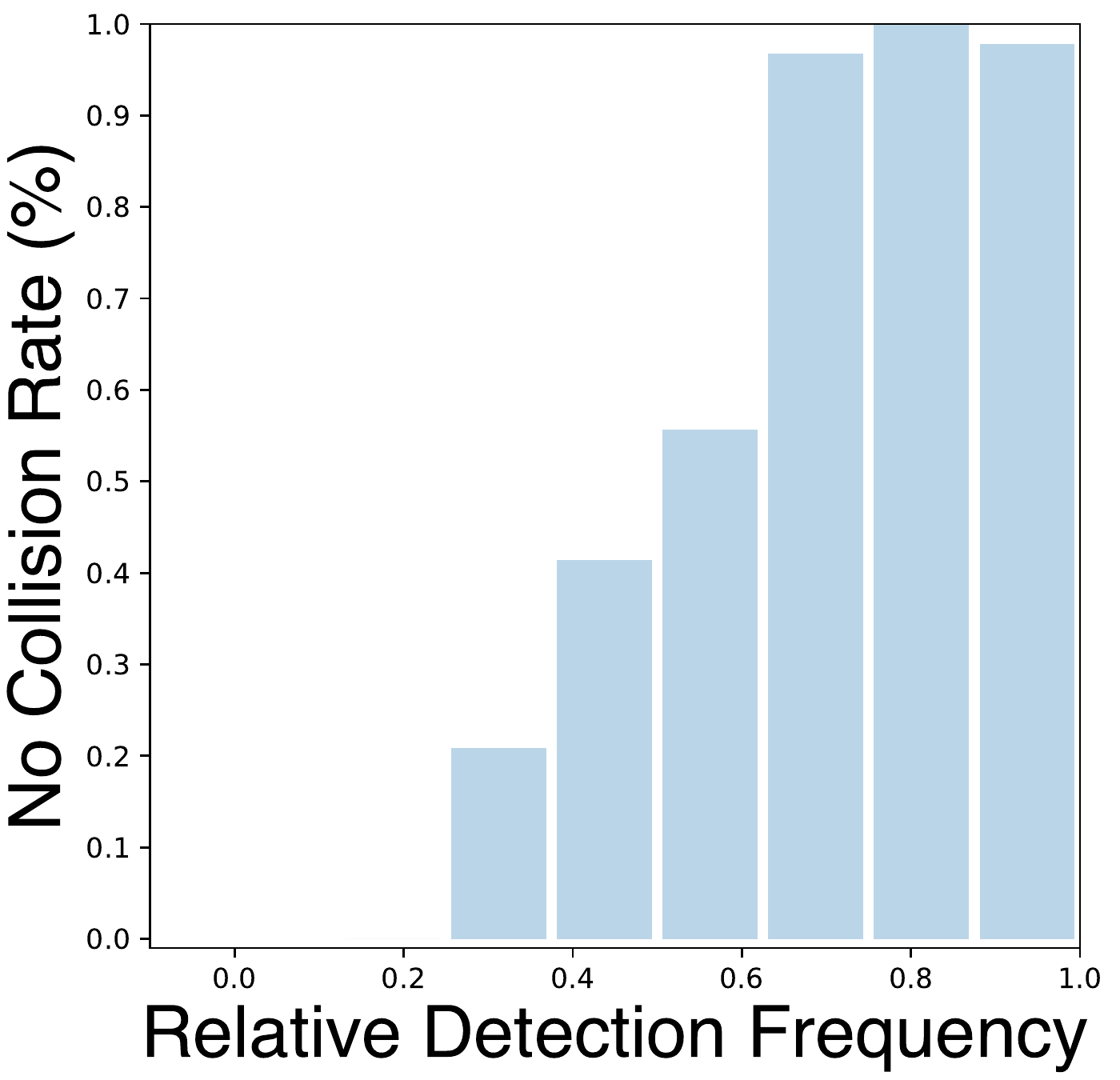}\label{fig:sp_c}}
      \hspace{0.2cm}
     \rulesep
     \hspace{0.1cm}
     \subfloat[][TC4]{\includegraphics[width=\sizescatter]{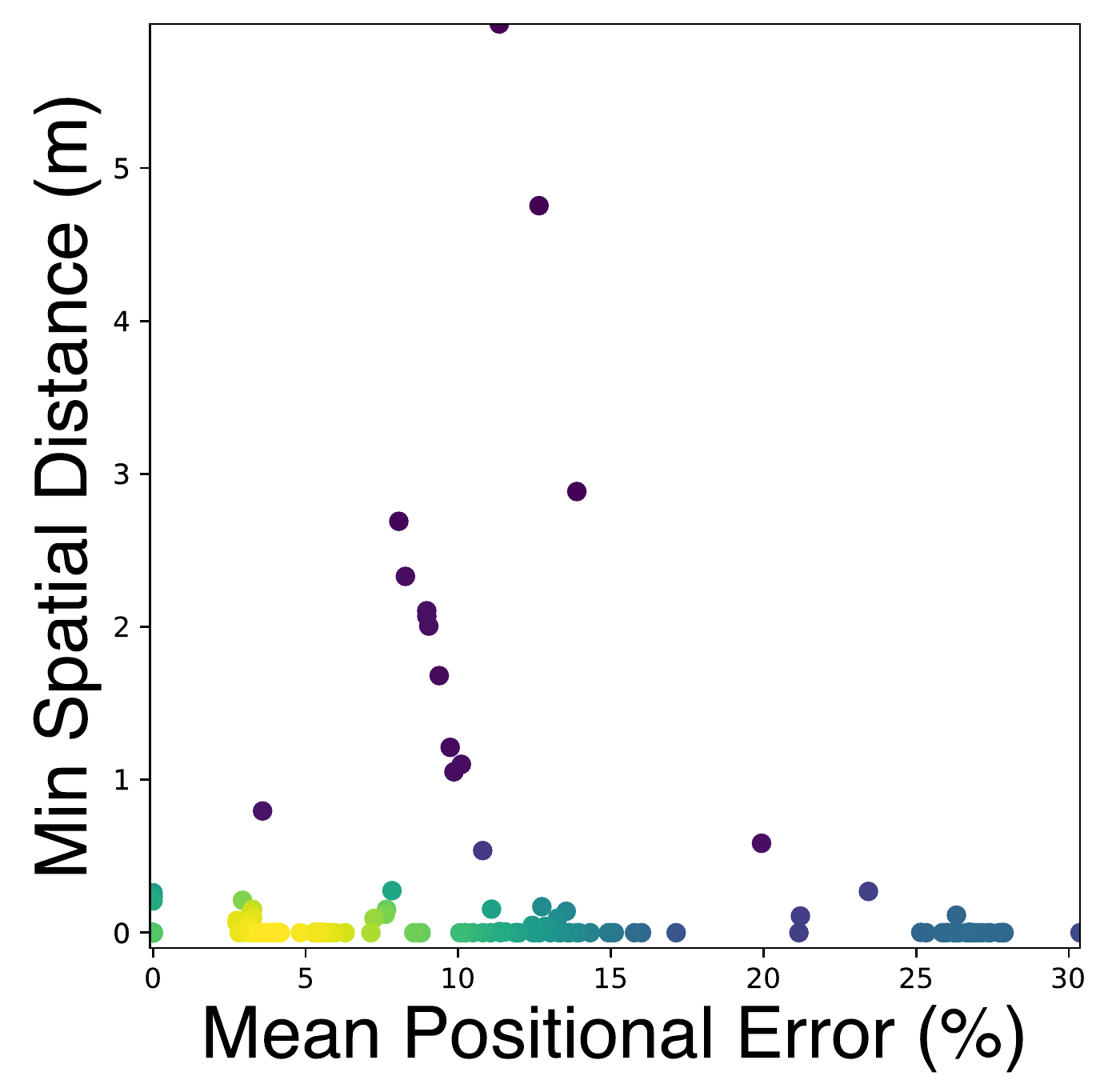}\label{fig:sp_d}}
     \hspace{0.2cm}
     \rulesep
     \hspace{0.1cm}
     \subfloat[][TC5b]{\includegraphics[width=\sizescatter]{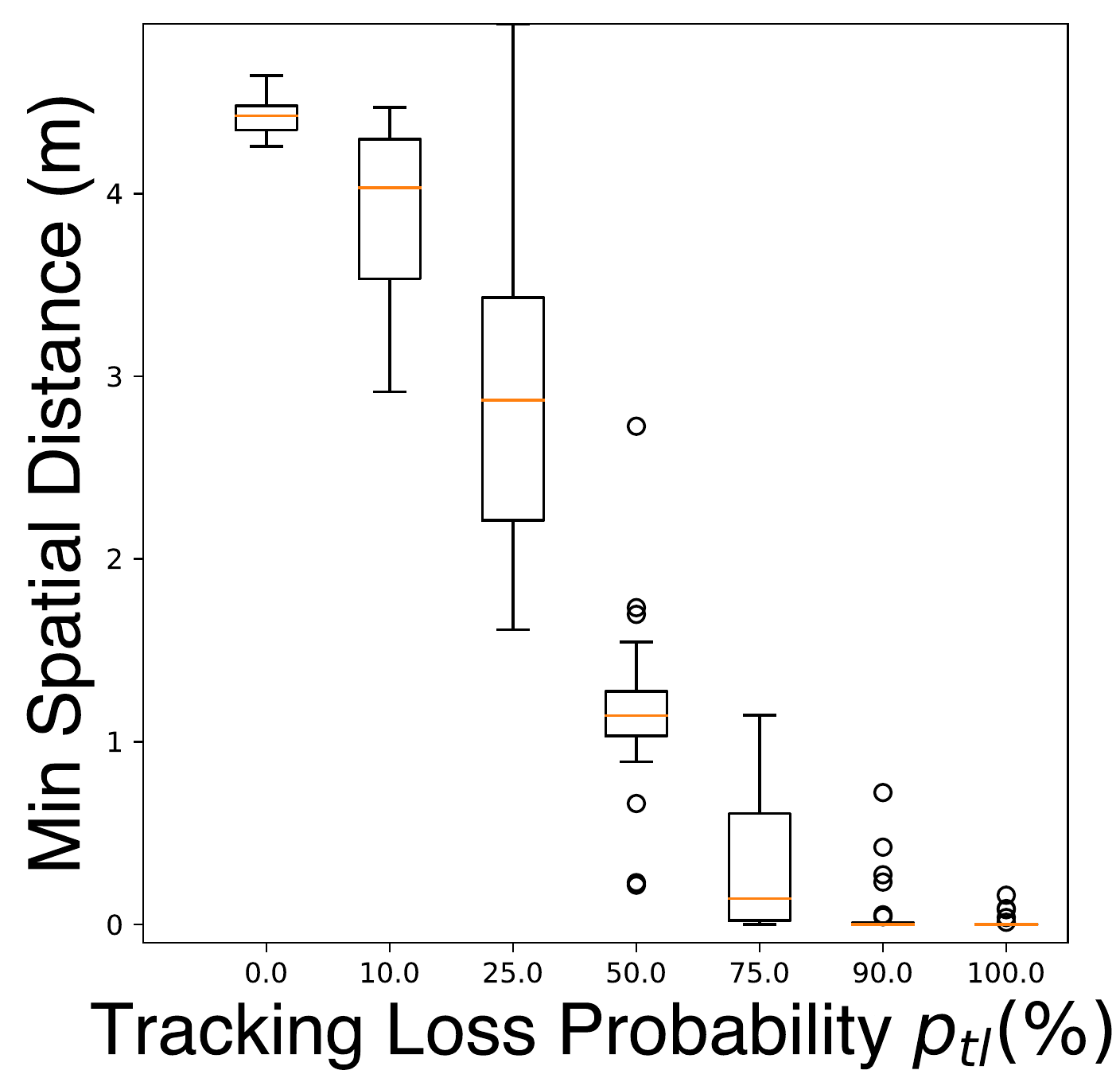}\label{fig:sp_e}}
     \\
     \vspace{-4mm} 
     \subfloat[][TC(1-3)]{\includegraphics[width=\sizescatter]{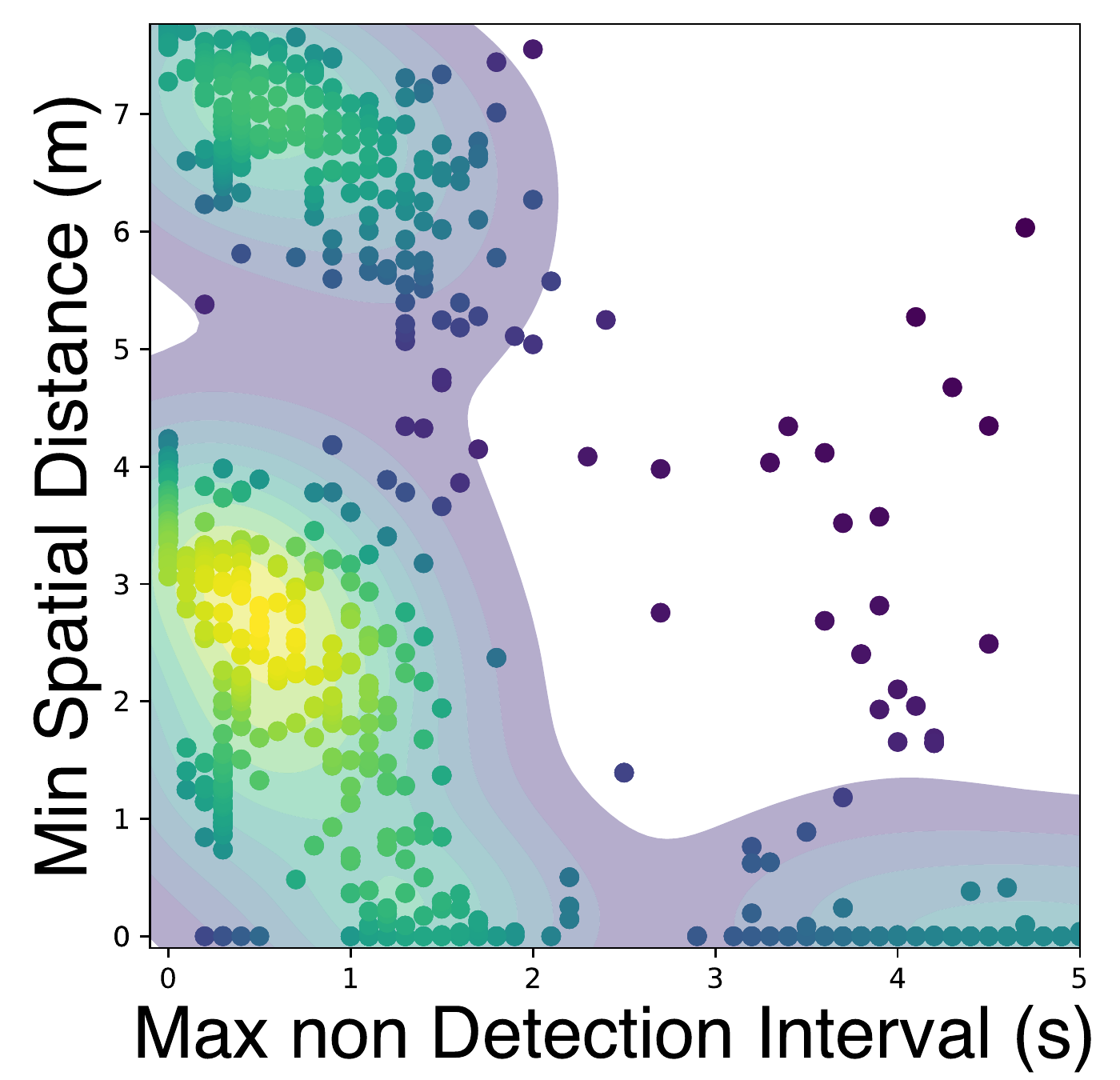}\label{fig:sp_f}}
     \hspace{1mm}
     \subfloat[][TC(1-3)]{\includegraphics[width=\sizescatter]{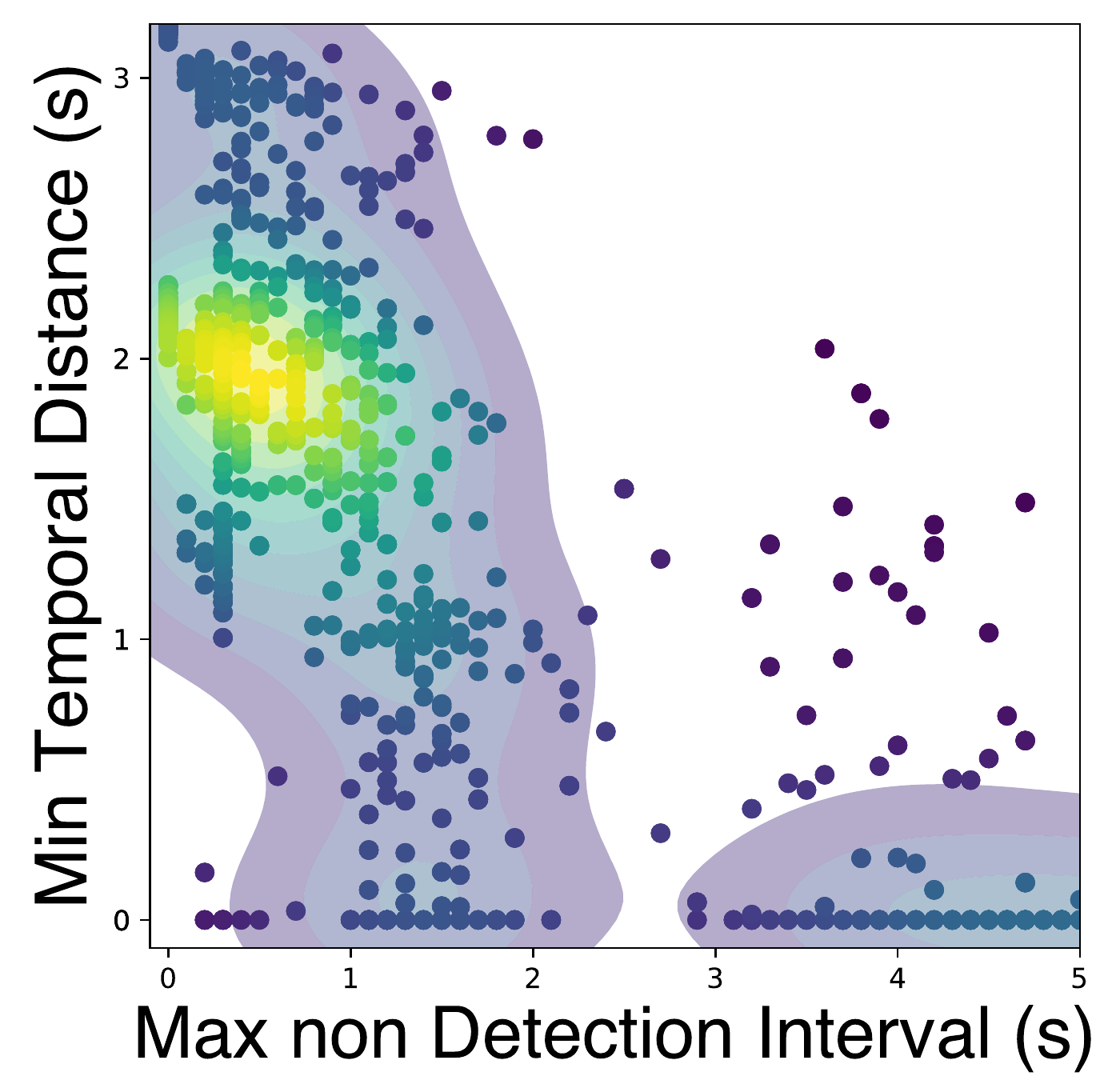}\textbf{}\label{fig:sp_g}}
     \hspace{1mm}
     \subfloat[][TC(1-3)]{\includegraphics[width=\sizescatter]{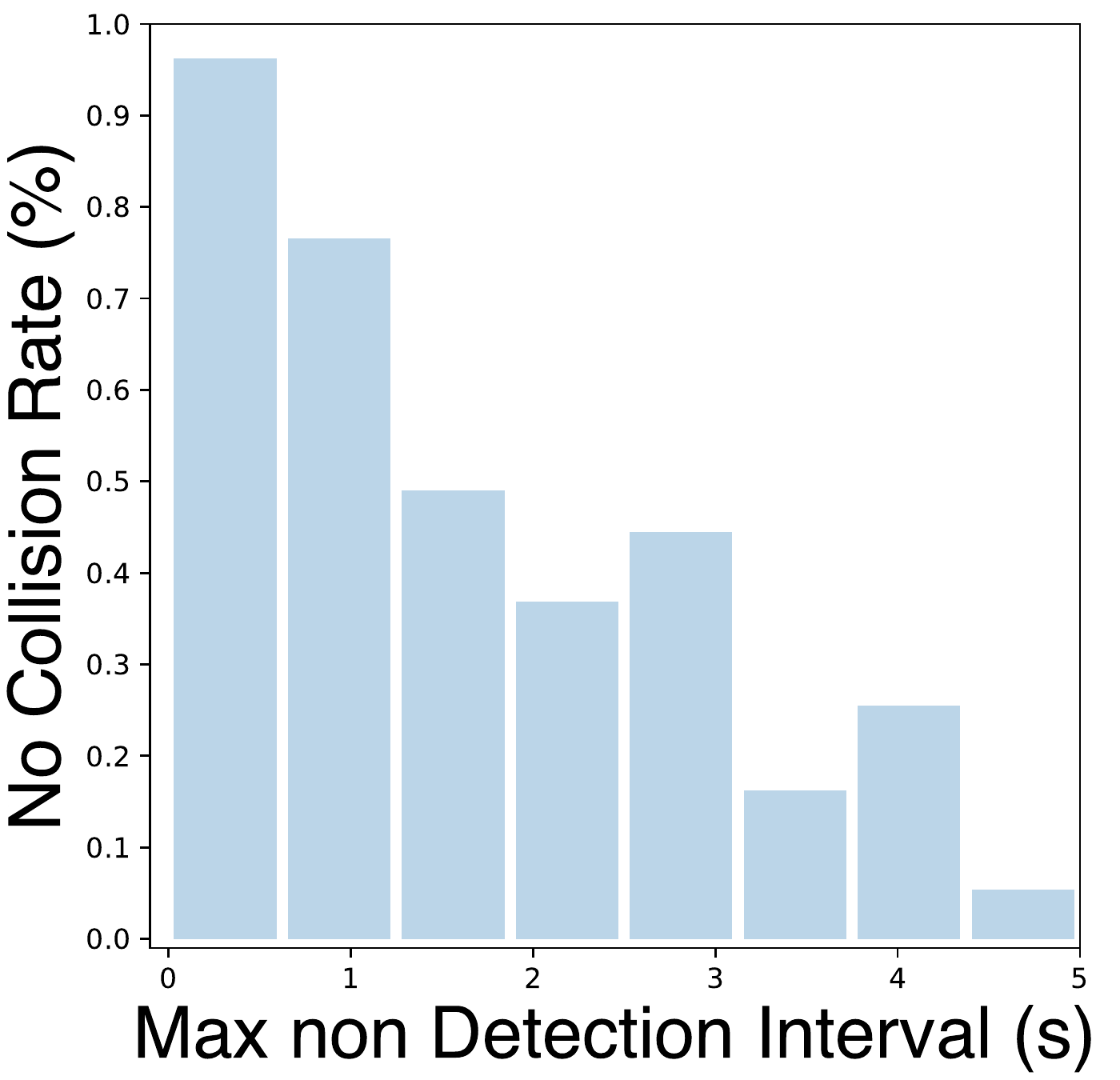}\label{fig:sp_h}}
      \hspace{0.2cm}
     \rulesep
     \hspace{0.1cm}
    \subfloat[][TC5a]{\includegraphics[width=\sizescatter]{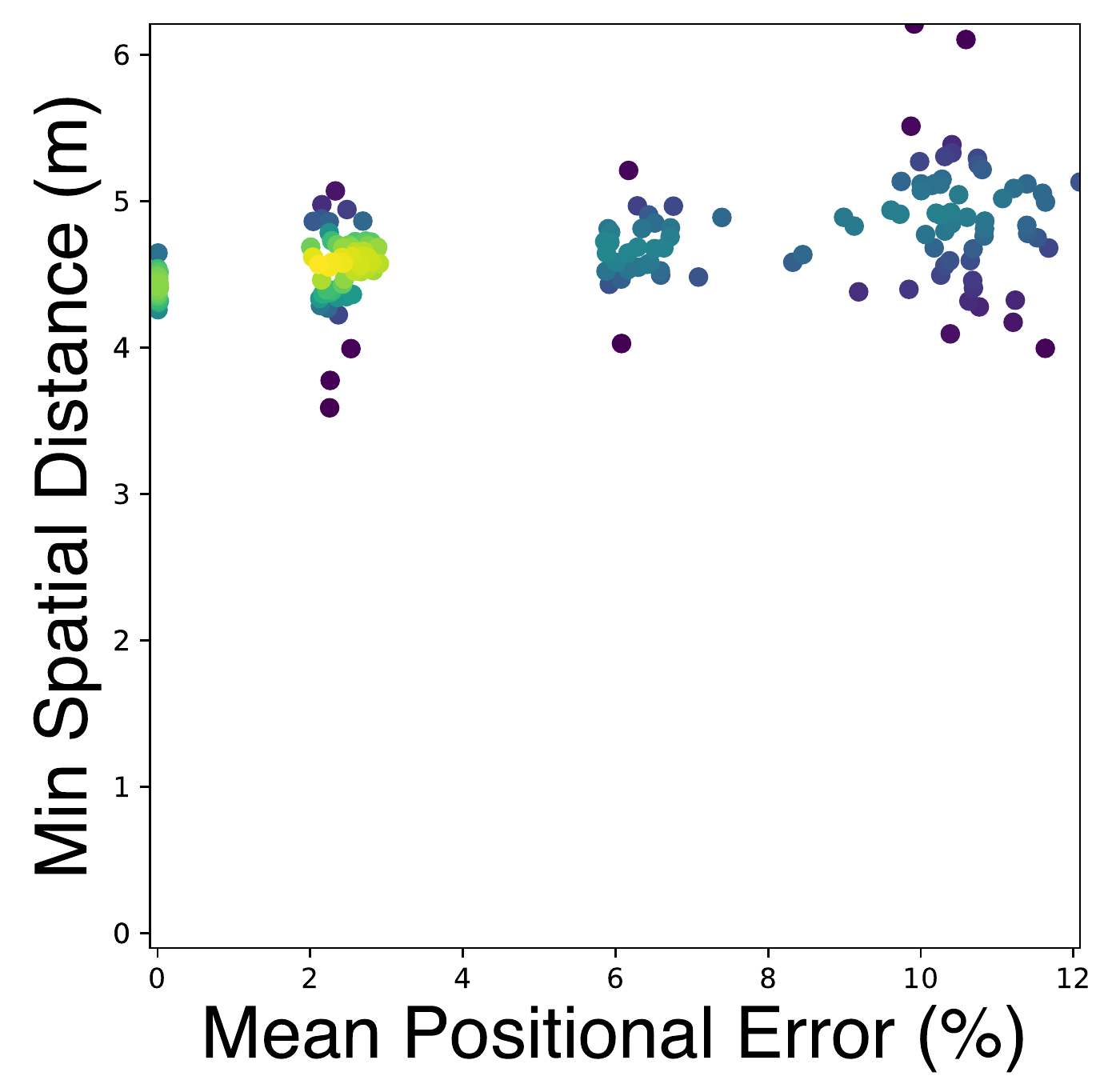}\label{fig:sp_i}}
    \hspace{0.2cm}
     \rulesep
     \hspace{0.1cm}
    \subfloat[][TC5b]{\includegraphics[width=\sizescatter]{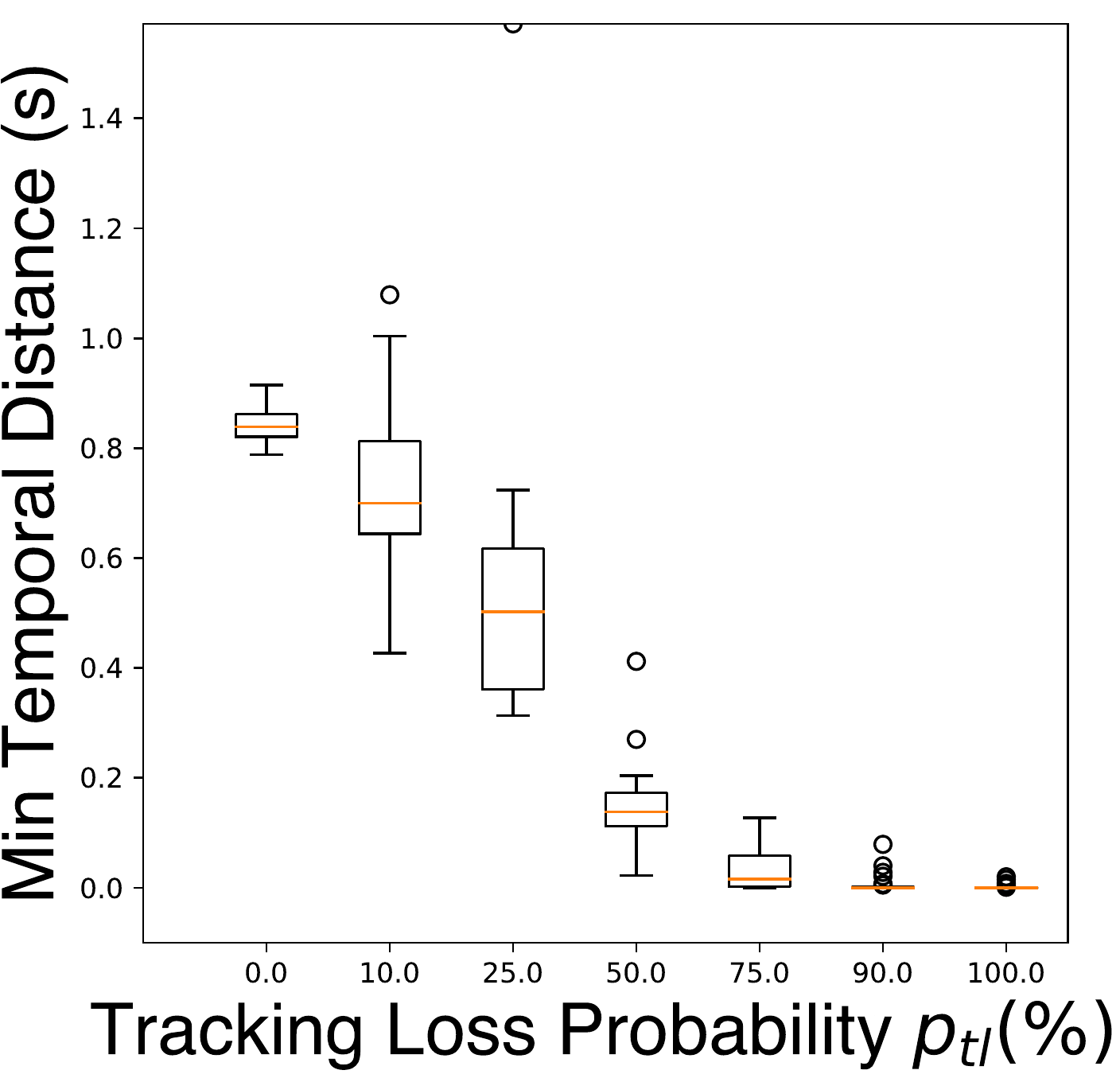}\label{fig:sp_j}}
     \vspace{-2mm} 
      \caption{Relationship between safety evaluation metrics and some specific statistics under varying PEMs. 
     These density scatter plots summarize all the runs of the relevant experiments, for specific test cases. Given the high number of samples (simulation runs), we highlighted the densest areas on a color scale from blue (low density) to yellow (high density).}
     \label{fig:scatterplots}
    \vspace{-3mm}
\end{figure*}

\begin{figure}[t]

     \centering
     \includegraphics[width=\columnwidth]{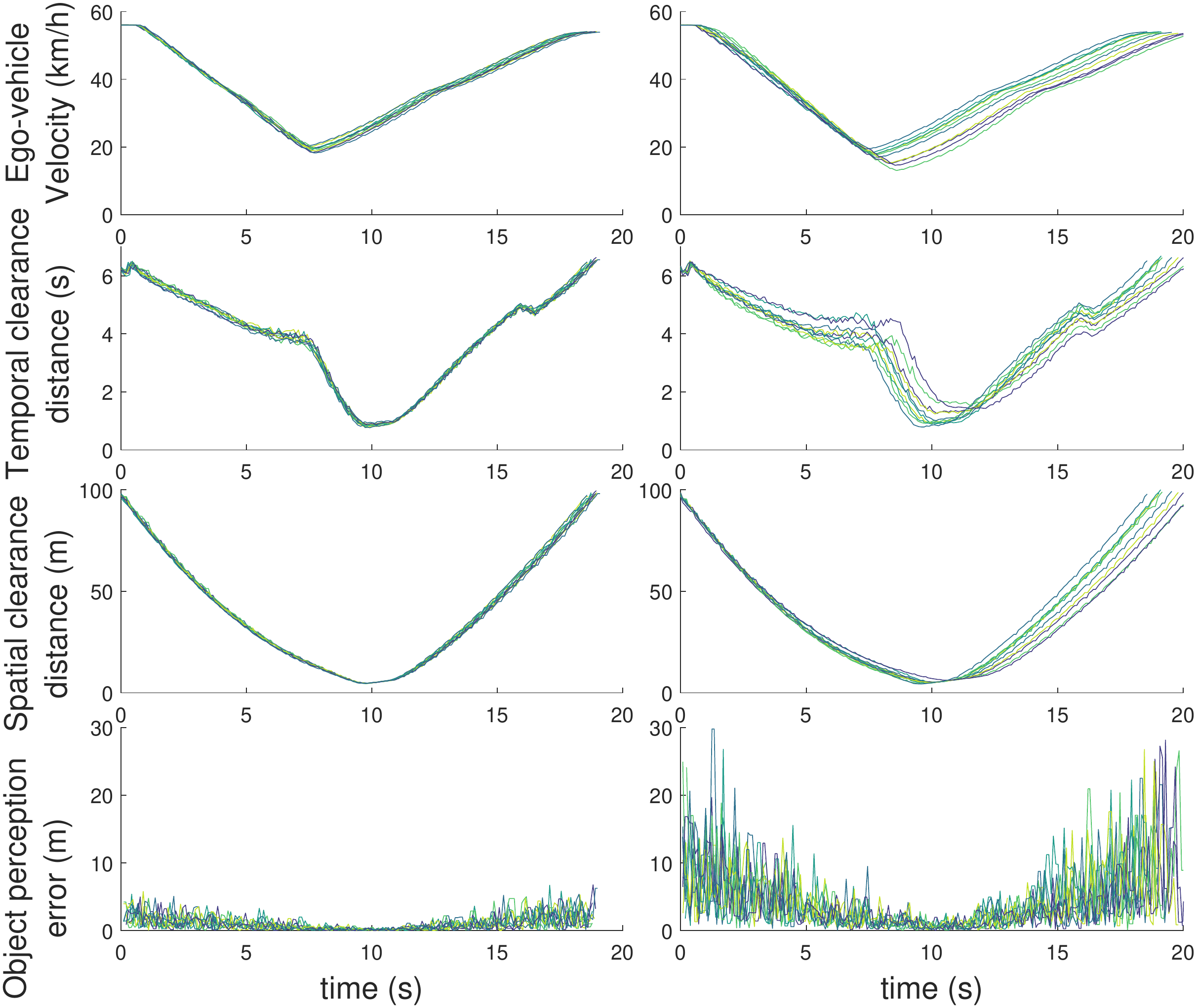}
     \caption{Illustration of ego-vehicle behavior for TC5a with two \PEMs: low $\sigma_d, \sigma_\theta$ (left), and high $\sigma_d, \sigma_\theta$ (right).
     }
     \label{fig:eval_illustration_VUT_behavior_errors_ped}
   
    \vspace{-2ex}
\end{figure}

\subsection{TC1-3: Following a traffic vehicle}
In Figures\autoref{fig:sp_a},\autoref{fig:sp_b},\autoref{fig:sp_f},\autoref{fig:sp_g}, we plot the relationship between two metrics for \textit{behavior evaluation}, namely minimum spatial distance (m) and minimum temporal distance (s), and two statistics of the perception error, namely, relative frequency of detection (realization of the \textit{steady state probability}) and maximum non-detection interval (realization of the \textit{mean sojourn time}).
In Figure\autoref{fig:sp_c}, we can observe that the success rate (no collision) is increasing with the increase of the relative detection frequency.
However, this is true up to a threshold of $\sim75\%$, above which the success rate is stationary.
 On the other hand, in Figure\autoref{fig:sp_h}, the duration of the non-detection intervals has a much more significant impact on the success rate.
Hence, we can observe that even in a situation of low visibility (i.e. low detection probability), if the intervals of non-detection are short enough, it is possible for the vehicle to avoid a collision.
This highlights the importance of including the temporal aspects of the error (critical issue I1) into the perception evaluation.

\subsection{TC4-5: Pedestrian on an urban road}
The second scenario offers a different insight. 
As illustrated in \mbox{Figures \autoref{fig:sp_d} and \autoref{fig:sp_i}}, we cannot observe major differences in safe behavior (minimum spatial clearance) under varying positional errors generated by different $PEM$s including the ground truth.
This indicates that the system failure is not due to the $\PEM$, but rather due to a weakness in the $DP$ of the ADS in \textit{our} experimental configuration,
which is unable to robustly handle TC4.
In fact, safety metrics in TC4 are not influenced by the magnitude of the positional error, since the system fails even with low/no errors. Similarly, in TC5a, the safety is not jeopardized by the error magnitude.
This also relates to the other two critical issues I2 and I3 of the current evaluation metrics.
Since a positional error by itself can still allow a safe response, it is not adequate to consider IoU as a metric for True Positive. 
On the contrary, a less restrictive metric should be considered, such as a distance threshold as proposed in \cite{nuscenes2019}.

In \autoref{fig:eval_illustration_VUT_behavior_errors_ped} we compare $\PEM$s with different positional errors in TC5a. Here also, the error magnitude does not have a major impact on safety, although smaller errors can lead to a more consistent AV behavior.
Furthermore, experiments for TC5b highlight how the safety decreases as $p_{tl}$ increases, as shown in \mbox{Figures \autoref{fig:sp_e} and \autoref{fig:sp_j}}. 
As $p_{tl}$ approaches $0.5$ and the obstacle velocity cannot be estimated, $DP$ is unable to predict the obstacle's trajectory and does not brake to avoid it, similar to the AV accident \cite{NTSB2019}.
This is in contrast to the findings in TC1-3, where frequent tracking errors were easier to handle than infrequent ones.
However, it provides an interesting insight towards understanding the contextual relevance of error types depending on the scenario.
In particular, in TC1-3 the obstacle is always in the path of the AV, while in cases such as TC5b their paths cross during the scenario (jaywalking in TC5b, but may be similarly applicable for a cut-in scenario).
In the latter case, proper obstacle trajectory prediction is critical to foresee the imminent collision.
\section{Conclusion}

In this paper, we have described an approach to test and study perception errors in a virtual environment,
by linking the respective performance of $S\&P$ and $DP$, 
and thereby enabling the identification of weaknesses in these subsystems. 
Furthermore, we have implemented an experimental setup to test handcrafted $\PEMs$ with the aim of highlighting some limitations of the currently used evaluation metrics for perception algorithms, while discussing how to analyze the resulting system behavior.
Although our focus is on AVs,
we believe that our approach is general enough to be applied to other domains involving navigational tasks and produce similar insights.

The main limitation of the current work lies in our focus on the detection of other road users.
Nevertheless, this is a key challenging problem hindering AV deployments, especially under adverse environmental conditions that degrade the $S\&P$ capabilities.
In the near future, we aim to further explore the study of \PEMs, develop more realistic simulations that incorporate perception errors, investigate the robustness of $S\&P$ under different environmental conditions, and finally, better approaches to test weaknesses of $DP$.

\clearpage

\bibliography{library}

\end{document}